\documentclass{article}

\usepackage{times,helvet,courier}
\usepackage[utf8]{inputenc}
\usepackage{amsmath}
\usepackage{amssymb}
\usepackage{microtype}
\usepackage{listings}
\usepackage{stmaryrd}
\usepackage{url}
\usepackage{xcolor}
\usepackage{ifthen}
\usepackage{wrapfig}

\usepackage{authblk}
\usepackage{amsthm}

\theoremstyle{definition}
\newtheorem{definition}{Definition}
\theoremstyle{plain}
\newtheorem{theorem}{Theorem}

\newtheorem{proposition}{Proposition}

\providecommand{\Underscore}{\textunderscore}

\lstdefinelanguage{clingo}{basicstyle=\ttfamily,keywordstyle=[1]\bfseries,keywordstyle=[2]\bfseries,keywordstyle=[3]\bfseries,showstringspaces=false,literate={_}{\Underscore}1 {\%\%}{}0,escapeinside={\#(}{\#)},alsoletter={\#,\&},keywords=[1]{not,from,import,def,if,else,elif,return,while,break,and,or,for,in,del,and,class,with,as,is,yield,async},keywords=[2]{\#const,\#show,\#minimize,\#base,\#theory,\#count,\#external,\#program,\#script,\#end,\#heuristic,\#edge,\#project,\#show,\#sum},morecomment=[l]{\#\ },morecomment=[l]{\%\ },morestring=[b]",stringstyle={\itshape},commentstyle={\color{darkgray}}}

\lstdefinelanguage{clingcon}[]{clingo}{morekeywords={&dom,&sum,&nsum,&diff,&disjoint,&distinct,&minimize,&maximize,&show}}
\lstdefinelanguage{fclingo}[]{clingo}{morekeywords={&sum,&sus,&in,&df,&min,&max,&show}}
\lstdefinelanguage{clingodl}[]{clingo}{morekeywords={&diff}}

\lstdefinelanguage{python}{basicstyle=\ttfamily,keywordstyle=[1]\bfseries,showstringspaces=false,literate={_}{\Underscore}{1},escapeinside={\#(}{\#)},alsoletter={\#,\&},keywords=[1]{not,from,import,def,if,else,elif,return,while,break,and,or,for,in,del,and,class,with,as,is,yield,async},morecomment=[l]{\#\ },morestring=[b]",stringstyle={\itshape},commentstyle={\color{darkgray}}}

 \providecommand{\sysfont}{\textit}

\newcommand{\Clingo}{\sysfont{Clingo}}

\newcommand{\clingcon}{\sysfont{clingcon}}
\newcommand{\clingo}{\sysfont{clingo}}
\newcommand{\clingodl}{\clingoM{dl}}

\newcommand{\telingo}{\sysfont{telingo}}

\newcommand{\clingoM}[1]{\clingo{\small\textnormal{[}\textsc{#1}\textnormal{]}}}

 \providecommand{\logfont}{\textrm}

\newcommand{\HT}{\ensuremath{\logfont{HT}}}

\newcommand{\HTC}{\ensuremath{\logfont{HT}_{\!c}}}

\newcommand{\MHT}{\ensuremath{\logfont{MHT}}}

 \RequirePackage{bm}
\RequirePackage{textcomp}
\RequirePackage{upgreek}

\newcommand{\next}{\text{\rm \raisebox{-.5pt}{\Large\textopenbullet}}}    

\newcommand{\alwaysF}{\ensuremath{\square}}

\newcommand{\eventuallyF}{\ensuremath{\Diamond}}

\newcommand{\finally}{\ensuremath{\bm{\mathsf{F}}}}
\newcommand{\initially}{\ensuremath{\bm{\mathsf{I}}}}

\mathchardef\mhyphen="2D

\newcommand{\intervco}[2]{\ensuremath{[#1..#2)}}

\newcommand{\rangeco}[3]{\ensuremath{#1 \in \intervco{#2}{#3}}}

\newcommand{\tuple}[1]{\ensuremath{\langle #1 \rangle}}
\newcommand{\Htrace}{\ensuremath{\mathbf{H}}}
\newcommand{\Ttrace}{\ensuremath{\mathbf{T}}}
\newcommand{\M}{\ensuremath{\mathbf{M}}}

\newcommand{\handt}{\tuple{H,T}}
\newcommand{\tandt}{\tuple{T,T}}

\lstdefinelanguage{clingos}{language=clingo,basicstyle=\small\ttfamily }
\makeatletter\lst@AddToHook{OnEmptyLine}{\vspace{\dimexpr-\baselineskip+\smallskipamount\relax}}\makeatother
\newcommand{\eqdef}{=} 

\providecommand{\Next}{\text{\rm \raisebox{-.5pt}{\Large\textopenbullet}}}  
\renewcommand{\initially}{\ensuremath{\mbox{$\bm{\mathsf{I}}$}}}
\newcommand{\metric}[3]{\ensuremath{#1_{
\ifthenelse{\equal{#2}{#3}}
{#2}
{
\ifthenelse{\equal{#2}{0}}
{
\ifthenelse{\equal{#3}{\omega}}
{}
{\leq#3}
      }
{
\ifthenelse{\equal{#3}{\omega}}
{\geq#2}
{\intervco{#2}{#3}}
      }
    }}}}
\newcommand{\metricI}[1]{\ensuremath{#1_{\cI}}}
\newcommand{\tmf}{\ensuremath{\tau}}
\newcommand{\cI}{\ensuremath{I}}

\newcommand{\x}{\ensuremath{x}}
\newcommand{\y}{\ensuremath{y}}
\newcommand{\melrule}[1]{\ensuremath{\alwaysF{(#1)}}}
\newcommand{\alphabet}{\ensuremath{\mathcal{A}}}
\newcommand{\program}{\ensuremath{P}}

\newcommand{\body}{\ensuremath{\beta}}
\newcommand{\head}{\ensuremath{\alpha}}

\newcommand{\allI}{\ensuremath{\mathbb{I}}}

\newcommand{\trans}[2]{(#1)_{#2}}
\newcommand{\tk}[1]{\trans{#1}{\kvar}}

\newcommand{\kvar}{\ensuremath{k}}
\newcommand{\tmvar}{\ensuremath{d}}
\newcommand{\timet}{\ensuremath{t}}

\newcommand{\tmflimit}{\ensuremath{\nu}}

\newcommand{\alphabetk}{\ensuremath{\mathcal{A}_k}}
\newcommand{\alphabets}{\ensuremath{\mathcal{A}^{*}}}

\newcommand{\alphabetT}{\ensuremath{\mathcal{T}}}

\newcommand{\den}[1]{\llbracket \, #1 \, \rrbracket}
\newcommand{\undefined}{\ensuremath{\boldsymbol{u}}} \newcommand{\Vh}{\ensuremath{h}} \newcommand{\Vt}{\ensuremath{t}} \newcommand{\htcinterp}{\ensuremath{\langle \Vh,\Vt \rangle}}
\newcommand{\htcttinterp}{\ensuremath{\langle \Vt,\Vt \rangle}}

\newcommand{\Labs}[3]{\ensuremath{{w_{#1}^{#2,#3}}}}
\newcommand{\Labsxy}[3]{\ensuremath{\Labs{\mu}{\x}{\y}}}

\newcommand{\mhtToHtf}{\ensuremath{\theta}} \newcommand{\mhtToHt}[1]{\ensuremath{\mhtToHtf(#1)}}
\newcommand{\mhtToHtc}[1]{\ensuremath{\mhtToHtf^c(#1)}}
\newcommand{\htToMhtf}{\ensuremath{\sigma}} \newcommand{\htToMht}[1]{\ensuremath{\htToMhtf(#1)}}
\newcommand{\htcToMht}[1]{\ensuremath{\htToMhtf^c(#1)}}

\newcommand{\true}{\ensuremath{\boldsymbol{t}}}
\newcommand{\false}{\ensuremath{\boldsymbol{f}}}

\renewcommand{\paragraph}[1]{\par\emph{#1}.\/}
  \newcommand{\paragraphC}[2]{\par\emph{#1}~\cite{#2}.}
\allowdisplaybreaks

\begin{document}

\title{Compiling Metric Temporal Answer Set Programming\thanks{Under consideration for publication in International Conference on Logic Programming and Nonmonotonic Reasoning (LPNMR 2024)}}

\author[1]{A. Becker}
\author[2]{P. Cabalar}
\author[3]{M. Diéguez}
\author[1]{J. Romero}
\author[1]{S. Hahn}
\author[1]{T. Schaub}

\affil[1]{University of Potsdam, Germany}
\affil[2]{University of Corunna, Spain}
\affil[3]{LERIA, University of Angers, France}
\date{}

\maketitle
\begin{abstract}
  We develop a computational approach to Metric Answer Set Programming (ASP)
  to allow for expressing quantitative temporal constrains, like durations and deadlines.
  A central challenge is to maintain scalability when dealing with fine-grained timing constraints,
  which can significantly exacerbate ASP's grounding bottleneck.
  To address this issue, we leverage extensions of ASP with difference constraints,
  a simplified form of linear constraints,
  to handle time-related aspects externally.
  Our approach effectively decouples metric ASP from the granularity of time,
  resulting in a solution that is unaffected by time precision.
\end{abstract}

 \section{Introduction}\label{sec:introduction}

Metric temporal logics~\cite{TIMEHandbook} allow for expressing quantitative temporal constrains,
like durations and deadlines.
As an example, consider the dentist scenario~\cite{mellarkod07a}:
\begin{wraptable}{r}{20mm}
  \begin{tabular}{|r|r|r|r|r|}
    \hline
               & \textit{D} & \textit{H} & \textit{O} & \textit{A} \\ \hline
    \textit{D} & 0          & 20         & 30         & 40         \\ \hline
    \textit{H} & 20         & 0          & 15         & 15         \\ \hline
    \textit{O} & 30         & 15         & 0          & 20         \\ \hline
    \textit{A} & 40         & 15         & 20         & 0          \\ \hline
  \end{tabular}
  \caption{Durations between locations} \label{table:dentist}
\end{wraptable}
{\em``Ram is at his office and has a dentist appointment in one hour.
For the appointment, he needs his insurance card which is at home and cash to pay the doctor.
He can get cash from the nearby Atm.
Table~\ref{table:dentist} shows the time in minutes needed to travel between locations: Dentist, Home, Office and Atm.
For example, the time needed to travel between Ram's office to the Atm is 20 minutes.
The available actions are: moving from one location to another and picking items such as cash or insurance.
The goal is to find a plan which takes Ram to the doctor on time.''}
This example combines planning and scheduling, and
nicely illustrates the necessity to combine qualitative and quantitative temporal constraints.

Extensions to the Logic of Here-and-There and Equilibrium Logic~\cite{pearce96a} were developed in~\cite{cadiscsc20a,becadiscsc23a}
to semantically ground the incorporation of metric constraints into Answer Set Programming~(ASP;~\cite{lifschitz19a}).
Building upon these semantic foundations, we develop a computational approach to metric ASP.
A central challenge in this is to maintain scalability when dealing with fine-grained timing constraints,
which can significantly exacerbate ASP's grounding bottleneck.
To address this issue, we leverage extensions of ASP with difference constraints~\cite{jakaosscscwa17a},
a simplified form of linear constraints,
to handle time-related aspects externally.
This approach effectively decouples metric ASP from the granularity of time,
resulting in a solution that is unaffected by time precision.
In detail, we
  (i) propose translations of metric logic programs into regular logic programs and their extension with difference constraints,
 (ii) prove the completeness and correctness of both translations in terms of equilibrium logic and its extensions
      with metric and difference constraints, and
(iii) describe an implementation of both translations in terms of meta encodings and
      give some indicative experimental results.
Conversely, we may consider metric logic programs as a high-level modeling language
for logic programs with difference constraints in the context of temporal domains.
We consider a simple yet expressive fragment in which the next operator can be supplied with a time frame expressed as an interval;
the full paper shows how this extends to the entire language.

 \paragraph{Related work}\label{sec:related:work}
Pioneering work on extending ASP with linear integer constraints to express quantitative temporal relations was done in~\cite{baboge05a,mellarkod07a}.
This work ultimately inspired the development of hybrid ASP systems such as \clingcon~\cite{bakaossc16a} and \clingodl~\cite{jakaosscscwa17a}.
Metric equilibrium logic was defined in~\cite{cadiscsc20a,becadiscsc23a} by building on Linear temporal equilibrium logic~\cite{agcadipescscvi20a}.
The latter provides the logical foundations of the temporal ASP system \telingo~\cite{cadilasc20a}.
Metric concepts are also present in stream reasoning, notably, the approach of \textit{lars}~\cite{bedaei18a}.
Metric extensions are also considered in Datalog~\cite{wagrkaka19a},
where they led to the \textit{meteor}~\cite{wahuwagr22a} and (temporal) \textit{vadalog}~\cite{beblnisa22a} systems.
Actions with durations in ASP are explored in~\cite{sobatu04a}.
Finally,
\cite{huozdi20a} considers reductions of (monotonic) Metric temporal logic to Linear temporal logic.
A comprehensive comparative account of metric approaches in logic programming is given in~\cite{becadiscsc23a}.
 \section{Background}
\label{sec:background}

To ease the formal elaboration of our translations from metric ASP to
regular logic programs and their extension with difference constraints,
we put ourselves into the context of the logical framework of Here-and-There and Equilibrium Logic.
\paragraphC{The Logic of Here-and-There}{pearce96a}
\label{sec:ht}
A formula over an alphabet \alphabet\ is defined as
\begin{align*}
  \varphi &::= \bot \mid a \mid \varphi\wedge\varphi \mid \varphi\vee\varphi \mid \varphi\to\varphi
  \quad\text{ where } a\in\alphabet\ .
\end{align*}
We define the derived operators
$\neg\varphi\eqdef\varphi\to\bot$
and
$\top\eqdef\neg\bot$.
Elements $a$ of \alphabet\ are called (Boolean) \emph{atoms}.
A \emph{literal} is an atom or an atom preceded by negation, viz.\ $a$ or $\neg a$.
A \emph{theory} is a set of formulas.
We sometimes write $\varphi\leftarrow\psi$ instead of $\psi\to\varphi$
to follow logic programming conventions.
A \emph{program} is a set of implications of the form $\varphi \leftarrow \psi$
where $\varphi$ is a disjunction of literals and $\psi$ is a conjunction of literals.

We represent an \emph{interpretation} $T$ as a set of atoms $T\subseteq\alphabet$.
An \HT-\emph{interpretation} is a pair $\tuple{H,T}$
of interpretations such that $H\subseteq T$;
it is said to be \emph{total} if $H=T$.
An HT-interpretation $\tuple{H,T}$ \emph{satisfies} a formula~$\varphi$,
written $\tuple{H,T}\models\varphi$,
if
\begin{enumerate}
\item $\tuple{H,T}\models a$ if $a \in H$
\item $\tuple{H,T}\models \varphi \wedge \psi$ if $\tuple{H,T} \models \varphi$ and $\tuple{H,T} \models \psi$
\item $\tuple{H,T}\models \varphi \vee   \psi$ if $\tuple{H,T} \models \varphi$ or  $\tuple{H,T} \models \psi$
\item $\tuple{H,T}\models \varphi \to    \psi$ if $\tuple{H',T} \not\models \varphi$ or $\tuple{H',T} \models \psi$
  for each~$H'\in\{H, T\}$
\end{enumerate}
An \HT-interpretation $\tuple{H,T}$ is an \emph{\HT-model} of a theory $\Gamma$
if $\tuple{H,T}\models\varphi$ for each~$\varphi\in\Gamma$.
A total model~$\tuple{T,T}$ of a theory~$\Gamma$ is an \emph{equilibrium model}
if there is no other model~$\tuple{H,T}$ of~$\Gamma$ with~$H \subset T$;
$T$ is also called a \emph{stable model} of $\Gamma$.
 \paragraphC{The Logic of Here-and-There with Constraints}{cakaossc16a}
\label{sec:htc}
The syntax of $\HTC$ relies on a signature~$\tuple{\mathcal{X},\mathcal{D},\mathcal{C}}$,
similar to constraint satisfaction problems,
where elements of set~$\mathcal{X}$ represent variables and elements of~$\mathcal{D}$ are domain values
(usually identified with their respective constants).
The constraint atoms in $\mathcal{C}$
provide an abstract way to relate values of variables and constants according to the atom's semantics.
For instance, \emph{difference constraint atoms} are expressions of the form `$x - y \leq d$',
containing variables~$x,y\in\mathcal{X}$ and the domain value~$d\in\mathcal{D}$.
A constraint formula $\varphi$ over $\mathcal{C}$ is defined as
\begin{align*}
  \varphi &::= \bot \mid c \mid \varphi\wedge\varphi \mid  \varphi\vee\varphi \mid  \varphi\rightarrow \varphi
  \quad\text{ where } c\in\mathcal{C}
\end{align*}
Concepts like defined operators, programs, theories, etc.\ are analogous to \HT.
Variables can be assigned some value from $\mathcal{D}$ or left \emph{undefined}.
For the latter, we use the special symbol $\undefined\not\in\mathcal{D}$ and
the extended domain $\mathcal{D}_u \eqdef \mathcal{D} \cup \{\undefined\}$.
A \emph{valuation} $v$ is a function $v:\mathcal{X}\rightarrow\mathcal{D}_u$.
We let $\mathbb{V}$ stand for the set of all valuations.
We sometimes represent a valuation $v$ as a set
\(
\{ (x,v(x)) \mid x\in\mathcal{X}, v(x)\in\mathcal{D}\}
\)
of pairs, so that $(x,\undefined)$ is never formed.
This allows us to use standard set inclusion, $v\subseteq v'$, for comparing $v,v'\in\mathbb{V}$.

The semantics of constraint atoms is defined in \HTC\ via \emph{denotations},
that is, functions
\(
\den{\cdot}:\mathcal{C}\rightarrow 2^{\mathbb{V}}
\)
mapping each constraint atom to a set of valuations.
An \emph{\HTC-interpretation} is a pair $\langle h,t \rangle$
of valuations such that $h\subseteq t$;
it is \emph{total} if $h=t$.
Given a denotation $\den{\cdot}$,
an \HTC-interpretation $\langle h,t \rangle$ \emph{satisfies} a constraint formula~$\varphi$,
written $\langle h,t \rangle \models \varphi$,
if
\begin{enumerate}
\item $\langle h,t \rangle \models c \text{ if } h\in \den{c}$
\item $\langle h,t\rangle \models \varphi\wedge\psi \text{ if } \langle h,t\rangle \models \varphi \text{ and } \langle h,t\rangle \models \psi$
\item $\langle h,t\rangle \models \varphi\vee\psi   \text{ if } \langle h,t\rangle \models \varphi \text{ or }  \langle h,t\rangle \models \psi$
\item $\langle h,t\rangle \models \varphi \rightarrow \psi
  \text{ if }\langle v,t\rangle \not\models \varphi
  \text{ or }\langle v,t\rangle     \models \psi
  \text{ for each }v\in\{h,t\}$
\end{enumerate}
An \HTC-interpretation~$\tuple{h,t}$ is an \emph{\HTC-model} of a theory~$\Gamma$
if $\tuple{h,t}\models\varphi$ for every~$\varphi\in\Gamma$.
A total model~$\langle t,t\rangle$ of a theory~$\Gamma$ is a \emph{constraint equilibrium model}
if there is no other model~$\tuple{h,t}$ of~$\Gamma$ with~$h\subset t$.
 \paragraphC{The Metric Temporal Logic of Here-and-There}{cadiscsc20a,becadiscsc23a}
\label{sec:mht}
Given an alphabet \alphabet\ and
\(
\allI\eqdef\{ \intervco{m}{n}\mid m\in\mathbb{N}, n\in\mathbb{N}\cup\{\omega\}\}
\),
a metric temporal formula is defined as\footnote{More general formulas, including until, release and past operators, are presented in~\cite{becadiscsc23a}.}
\begin{align*}
  \varphi &::= \bot \mid a \mid \varphi\wedge\varphi \mid \varphi\vee\varphi \mid \varphi\to\varphi \mid
  \initially                     \mid
  \metricI{\next}        \varphi \mid
  \metricI{\alwaysF}     \varphi \mid
  \metricI{\eventuallyF} \varphi
  \quad\text{ where } a\in\alphabet, I\in\allI
\end{align*}
The last three cases deal with metric temporal operators,
which are indexed by some interval $I$.
In words,
\metricI{\next},
\metricI{\alwaysF}, and
\metricI{\eventuallyF}
are called
\emph{next},
\emph{always}, and
\emph{eventually}.
\initially\ simply refers to the initial state.
We write
$\next$, $\alwaysF$, $\eventuallyF$ for
$\next_{\intervco{0}{\omega}}$, $\alwaysF_{\intervco{0}{\omega}}$, $\eventuallyF_{\intervco{0}{\omega}}$, respectively.
In addition to the aforedefined Boolean operators,
we define
$\finally   \eqdef  \neg \metric{\next}{0}{\omega}     \top$,
which allow us to refer to the final state. Concepts like programs, theories, etc.\ are analogous to \HT.

The semantics of temporal formulas is defined via \emph{traces},
being sequences $\Ttrace=(T_i)_{\rangeco{i}{0}{\lambda}}$ of interpretations $T_i\subseteq\alphabet$;
$\lambda$ is the length of \Ttrace.
Here, we consider only traces of finite length. We define the ordering $\Htrace\leq\Ttrace$ between traces of the same length $\lambda$ as $H_i\subseteq T_i$ for each $\rangeco{i}{0}{\lambda}$,
and $\Htrace<\Ttrace$ as both $\Htrace\leq\Ttrace$ and $\Htrace\neq\Ttrace$.
An \emph{\HT-trace} over \alphabet\ of length $\lambda$ is
a sequence of pairs
\(
(\tuple{H_i,T_i})_{\rangeco{i}{0}{\lambda}}
\)
with $H_i\subseteq T_i$ for any $0 \leq i < \lambda$;
it is \emph{total} if $H_i=T_i$.
For convenience, we represent it as the pair $\tuple{\Htrace,\Ttrace}$ of traces
$\Htrace = (H_i)_{\rangeco{i}{0}{\lambda}}$ and $\Ttrace = (T_i)_{\rangeco{i}{0}{\lambda}}$.

Metric information is captured by timing functions.
Given $\lambda\in\mathbb{N}$,
we say that
\(
\tmf: \intervco{0}{\lambda} \to \mathbb{N}
\)
is a (strict) \emph{timing function} wrt $\lambda$
if
$\tmf(0)=0$ and
$\tmf(\kvar)<\tmf(\kvar+1)$ for $0\leq\kvar<\lambda-1$.
A \emph{timed} \HT-trace $(\tuple{\Htrace,\Ttrace},\tmf)$ over \alphabet\ and $(\mathbb{N},<)$ of length $\lambda$
is a pair consisting of
an \HT-trace $\tuple{\Htrace,\Ttrace}$ over \alphabet\ of length $\lambda$ and
a timing function $\tmf$ wrt $\lambda$.
A timed \HT-trace $\M=(\tuple{\Htrace,\Ttrace}, \tmf)$
of length $\lambda$ over alphabet \alphabet\
\emph{satisfies} a metric formula $\varphi$ at $\rangeco{k}{0}{\lambda}$,
written \mbox{$\M,k \models \varphi$}, if
\begin{enumerate}
\item $\M,k \models a$ if $a \in H_k$
\item \label{def:mhtsat:and} $\M, k \models \varphi \wedge \psi$
  if
  $\M, k \models \varphi$
  and
  $\M, k \models \psi$
\item \label{def:mhtsat:or} $\M, k \models \varphi \vee \psi$
  if
  $\M, k \models \varphi$
  or
  $\M, k \models \psi$
\item $\M, k \models \varphi \to \psi$
  if
  $\M', k \not \models \varphi$
  or
  $\M', k \models  \psi$,
  \\ for both $\M'=\M$ and $\M'=(\tuple{\Ttrace,\Ttrace}, \tmf)$
\item $\M, k \models \initially$
  if
  $k =0$
\item \label{def:mhtsat:next}$\M, k \models \metricI{\next}\, \varphi$
  if
  $k+1<\lambda$ and $\M, k{+}1 \models \varphi$
  and $\tmf(k{+}1)-\tmf(k) \in \cI$
\item \label{def:mhtsat:eventuallyF} $\M, k \models \metricI{\eventuallyF}\, \varphi$
  if
  $\M, i \models \varphi$ for some $\rangeco{i}{k}{\lambda}$
  with
  $\tmf(i)-\tmf(k) \in \cI$
\item \label{def:mhtsat:alwaysF} $\M, k \models \metricI{\alwaysF}\, \varphi$
  if
  $\M, i \models \varphi$ for all $\rangeco{i}{k}{\lambda}$
  with
  $\tmf(i)-\tmf(k) \in \cI$
\end{enumerate}
A timed \HT-trace $\M$ is an \emph{\MHT-model} of a metric theory $\Gamma$ if $\M,0 \models \varphi$ for all $\varphi \in \Gamma$.
A total \MHT-model~$(\tuple{\Ttrace, \Ttrace}, \tmf)$ of a theory~$\Gamma$ is a \emph{metric equilibrium model}
if there is no other model $(\tuple{\Htrace,\Ttrace}, \tmf)$ of~$\Gamma$ with $\Htrace<\Ttrace$.
 \par
\newcommand{\distloc}{\ensuremath{\delta\langle L,L'\rangle}}
\newcommand{\tat}{\textit{at}}
\newcommand{\tram}{\textit{ram}}
\newcommand{\tgo}{\textit{go}}
\newcommand{\thas}{\textit{has}}
\newcommand{\toffice}{\textit{office}}
\newcommand{\thome}{\textit{home}}
\newcommand{\tdentist}{\textit{dentist}}
\newcommand{\tatm}{\textit{atm}}
\newcommand{\tcard}{\textit{card}}
\newcommand{\tcash}{\textit{cash}}
\newcommand{\ticard}{\textit{icard}}
\newcommand{\tgoal}{\textit{goal}}
For illustration,
let us consider the formalization of the dentist scenario in~\eqref{ex:dentist:one} to~\eqref{ex:dentist:ten}.
We assume that variables $L$ and $L'$ are substituted by distinct locations $\toffice$, $\tatm$, $\tdentist$, and $\thome$;
and variable $I$ by items $\tcash$ and $\ticard$.
We use \distloc\ to refer to the distance between two locations from Table~\ref{table:dentist}.
As in \cite{mellarkod07a},
we assume that Ram is automatically picking up items when being at the same position. \begin{align}
\alwaysF(\tat(\tram,\toffice)                                   &\leftarrow \initially)\label{ex:dentist:one}\\
  \alwaysF(\tat(\tcash,\tatm)                                     &\leftarrow \initially)\\
  \alwaysF(\tat(\ticard,\thome)                                   &\leftarrow \initially)\label{ex:dentist:tri}\\
\alwaysF(\textstyle\bigvee_{L'\neq L} \tgo(\tram,L')            &\leftarrow \tat(\tram,L)\wedge \neg\finally)\label{ex:dentist:for}\\
\alwaysF(\thas(\tram,I)                                         &\leftarrow \tat(\tram,L) \wedge \tat(I,L))\label{ex:dentist:six}\\
  \alwaysF(\tat(I,L)                                              &\leftarrow \tat(\tram,L) \wedge \thas(\tram,I))\label{ex:dentist:svn}\\
\alwaysF(\Next_{\intervco{\distloc}{\distloc+1}}{\tat(\tram,L')}&\leftarrow\tat(\tram,L) \wedge \tgo(\tram,L'))\label{ex:dentist:eit}\\
\alwaysF(\Next{\thas(\tram,I)}                                  &\leftarrow \thas(\tram,I) \wedge \neg\finally)\label{ex:dentist:nin}\\
  \alwaysF(\Next{\tat(I,L)}                                       &\leftarrow \neg \thas(\tram,I) \wedge \tat(I,L) \wedge \neg\finally)\label{ex:dentist:ten}
\end{align}
In brief,
Rules~\eqref{ex:dentist:one} to~\eqref{ex:dentist:tri} give the initial situation.
Rule~\eqref{ex:dentist:for} delineates possible actions.
Rules~\eqref{ex:dentist:six} and~\eqref{ex:dentist:svn} capture indirect effects.
Rule~\eqref{ex:dentist:eit} is the effect of moving from location $L$ to $L'$;
it uses the next operator restricted by the duration between locations.
Rules~\eqref{ex:dentist:nin} and~\eqref{ex:dentist:ten} address inertia.

 \section{Metric Logic Programs}

Metric logic programs,
defined as metric theories composed of implications akin to logic programming rules,
derive their semantics from their metric equilibrium models.
Syntactically,
a \emph{metric logic program}
over \alphabet\
is a set of \emph{metric rules} of form
\[
  \melrule{\head\leftarrow\body}
  \quad\text{ or }\quad
  \melrule{\metricI{\Next} a\leftarrow\body}
\]
for
$\head\eqdef a_1\vee  \dots\vee   a_m\vee  \neg a_{m+1}\vee  \dots\vee  \neg a_n$,
$\body\eqdef b_1\wedge\dots\wedge b_o\wedge\neg b_{o+1}\wedge\dots\wedge\neg b_p$, and
$m,n,o,p\in\mathbb{N}$
with
$a,a_i\in\alphabet$
for
$1\leq i\leq n$,
and
$b_i\in\alphabet\cup\{\initially,\finally\}$
for
$1\leq i\leq p$.

While our considered language fragment excludes global temporal operators and disjunctive metric heads,
it effectively captures state transitions and allows for imposing timing constraints upon them.
A comprehensive treatment is provided in our full paper,
but they are omitted here for simplicity as they require more elaborate translations.

Our two alternative translations share a common structure, each divided into three distinct parts.
The first part maps a metric program into a regular one.
This part captures the state transitions along an \HT-trace specified by the metric program,
and is common to both translations.
The second and third part capture the timing function along with
its interplay with the interval constraints imposed by the metric program, respectively.
The two variants of these parts are described in Section~\ref{sec:mlp:ht} and~\ref{sec:mlp:htc} below.

The first part of our translation takes a metric program over \alphabet\
and yields rules over
\(
\alphabets\eqdef\bigcup\nolimits_{\kvar\in\mathbb{N}} \alphabetk
\)
for
\(
\alphabetk\eqdef\{a_\kvar \mid a \in \alphabet \}
\)
and $\kvar \in \mathbb{N}$.
Atoms of form $a_k$ in $\alphabets$ represent the values taken by variable $a\in\alphabet$
at different points $k$ along a trace of length $\lambda$.

We begin by inductively defining the translation $\tk{r}$ of a metric rule $\alwaysF{r}$ at $k$
for $0 \leq \kvar < \lambda$ and $\lambda\in\mathbb{N}$
as follows:\footnote{Note that $\top$, $\neg$ and $\finally$ are defined operators.}
\par
\begin{minipage}[t]{0.4\linewidth}\vspace{-15pt}\begin{align*}
  \tk{a}                &\eqdef a_\kvar \text{ if } a \in \alphabet \\
  \tk{\metricI{\next}a} &\eqdef
                          \begin{cases}
                            \bot    & \text{if } \kvar = \lambda-1 \\[-1pt]
                            \trans{a}{\kvar+1}        & \text{otherwise}
                          \end{cases} \\
  \tk{\initially}       &\eqdef
                          \begin{cases}
                            \top        & \text{if } \kvar = 0 \\[-3pt]
                            \bot        & \text{otherwise}
                          \end{cases} \end{align*}\vspace{-5pt}\end{minipage}
\qquad
\begin{minipage}[t]{0.4\linewidth}\vspace{-15pt}\begin{align*}
\tk{\bot}                    &\eqdef \bot\\[1pt]
\tk{\varphi_1\wedge\varphi_2}&\eqdef \tk{\varphi_1}\wedge\tk{\varphi_2}  \\[1pt]
  \tk{\varphi_1\vee\varphi_2}  &\eqdef \tk{\varphi_1}\vee\tk{\varphi_2}    \\[1pt]
  \tk{\head \leftarrow\body}   &\eqdef\{\tk{\varphi_1}\leftarrow\tk{\varphi_2}\}
\end{align*}\vspace{-5pt}\end{minipage}
\\
Note that we drop the always operator \alwaysF{} preceding metric rules in the translation;
it is captured by producing a rule instance for every $0 \leq \kvar < \lambda$.
Accordingly, for a metric program \program\ over \alphabet\ and $\lambda\in\mathbb{N}$, we define
\begin{align*}
  \Pi_\lambda(\program)
  &=
  \textstyle\bigcup_{\alwaysF{r} \in P,\, 0 \leq \kvar < \lambda} \tk{r}
  \quad\text{ over }
  \alphabets\ .
\end{align*}

For illustration,
consider the instance of \eqref{ex:dentist:eit} for moving from $\toffice$ to $\thome$
\begin{align}\label{ex:dentist:eit:translated}
  \alwaysF(\Next_{\intervco{15}{16}}{\tat(\tram,\thome)}&\leftarrow\tat(\tram,\toffice) \wedge \tgo(\tram,\thome)) \ .
\end{align}
Our translation ignores \alwaysF\ at first and yields:
\begin{align}
                      \bot&\leftarrow\tat(\tram,\toffice)_k \wedge \tgo(\tram,\thome)_k\quad\text{ for }\kvar = \lambda-1\label{ex:dentist:eit:translated:one}\\
  \tat(\tram,\thome)_{k+1}&\leftarrow\tat(\tram,\toffice)_k \wedge \tgo(\tram,\thome)_k\quad\text{ otherwise}\label{ex:dentist:eit:translated:two}
\end{align}
When assembling $\Pi_\lambda(\program)$ for $\lambda=100$ and $P$ being the rules in~\eqref{ex:dentist:one}
to~\eqref{ex:dentist:ten}, we account for \alwaysF\ by adding 99 instances of the rule
in~\eqref{ex:dentist:eit:translated:two}
and a single instance of~\eqref{ex:dentist:eit:translated:one}.

This first part of our translation follows Kamp's translation~\cite{kamp68a} for Linear temporal logic.
Of particular interest is the translation of $\metricI{\Next} a\leftarrow\body$.
The case analysis accounts for the actual state transition of the next operator,
which is infeasible at the end of the trace.
Thus, we either derive $a_{k+1}$ or a contradiction.
The metric aspect is captured by the translations in Section~\ref{sec:mlp:ht} and~\ref{sec:mlp:htc}.
Whenever all intervals in a metric programs $P$ are of form \intervco{0}{\omega},
we get a one-to-one correspondence between
\MHT-traces of length $\lambda$ of $P$ with an arbitrary yet fixed timing function and
\HT-interpretations of $\Pi_\lambda(\program)$.
Finally, it is worth noting that the size of the resulting program $\Pi_\lambda(\program)$ grows with $\lambda$.
 \section{Translation of Metric Logic Programs to \HT}
\label{sec:mlp:ht}

We begin by formalizing timing functions $\tau$ via Boolean atoms in
\(
\alphabetT\eqdef\{\timet_{\kvar,\tmvar} \mid \kvar,\tmvar\in\mathbb{N} \}.
\)
An atom like $\timet_{\kvar,\tmvar}$ represents that $\tau(\kvar)=\tmvar$.
To obtain finite theories,
we furthermore impose an upper bound $\tmflimit\in\mathbb{N}$ on the range of $\tau$.
Hence, together with the trace length $\lambda$,
our formalization $\Delta_{\lambda,\tmflimit}$ only captures
timing functions $\tau$ satisfying $\tmf(\lambda-1)\leq\tmflimit$.

Given $\lambda,\tmflimit \in \mathbb{N}$,
we let
\begin{align}\label{def:ht:delta}
  \Delta_{\lambda,\tmflimit} &\eqdef
     \{ \timet_{0,0} \} \cup
     \{ \textstyle\bigvee_{d<d'\leq\tmflimit}\timet_{\kvar+1,\tmvar'}\leftarrow\timet_{\kvar,\tmvar} \mid
            0 \leq \kvar < \lambda-1, 0 \leq \tmvar \leq \tmflimit\}
\end{align}
Starting from $\tau(0)=0$, represented by $\timet_{0,0}$,
the rule in~\eqref{def:ht:delta} assigns strictly increasing time points to consecutive states,
reflecting that $\tmf(\kvar)<\tmf(\kvar+1)$ for $0\leq\kvar<\lambda-1$.

The last part of our formalization accounts for the interplay of the timing function with the interval conditions imposed in the program.
Given $\lambda,\tmflimit \in \mathbb{N}$
and
a metric program $\program$,
we let
\begin{align}
  \Psi_{\lambda,\tmflimit}(\program) \eqdef
  &\ \{ \bot \leftarrow \tk{\body} \wedge \timet_{\kvar,\tmvar} \wedge \timet_{\kvar+1,\tmvar'} \mid
    0 \leq \kvar < \lambda-1, 0\leq d<d'\leq\tmflimit,
    \label{def:ht:psi:one}\\&\qquad\qquad\nonumber
  \phantom{n\in\mathbb{N}, }\tmvar'-\tmvar < m, \melrule{\metric{\next}{m}{n} a \leftarrow \body} \in \program \}\;\cup \\
  &\ \{ \bot \leftarrow \tk{\body} \wedge \timet_{\kvar,\tmvar} \wedge \timet_{\kvar+1,\tmvar'} \mid
    0 \leq \kvar < \lambda-1, 0\leq d<d'\leq\tmflimit,
    \label{def:ht:psi:two}\\&\qquad\qquad\nonumber
           n\in\mathbb{N},  \tmvar'-\tmvar \geq n, \melrule{\metric{\next}{m}{n} a \leftarrow \body} \in \program \}
\end{align}
The integrity constraints ensure that for every metric rule $\melrule{\metric{\next}{m}{n} a \leftarrow \body}$
the duration between the $k$th and $(k+1)$st state in a trace
falls within interval $\intervco{m}{n}$.
With $\tmf(\kvar) =d$ and $\tmf(\kvar+1) =d'$,
this amounts to checking whether
\(
\tmvar+m \leq \tmvar' < \tmvar+n
\),
if $n$ is finite;
otherwise, the verification of the upper bound in~\eqref{def:ht:psi:two} is dropped for $n=\omega$.

Note that the size of both $\Delta_{\lambda,\tmflimit}$ and $\Psi_{\lambda,\tmflimit}(\program)$
is proportional to $\mathcal{O}(\lambda\cdot\tmflimit^2)$.
Hence, long traces and even more severely fine-grained timing functions lead to a significant blow up
when translating metric programs into regular ones with the above formalization.

For the rule in~\eqref{ex:dentist:eit:translated},
we get
\begin{align}
  \bot&\leftarrow\tat(\tram,\toffice)_k\wedge\tgo(\tram,\thome)_k\wedge\timet_{\kvar,\tmvar}\wedge\timet_{\kvar+1,\tmvar'}
        \quad\text{ for } d'-d < 15
  \\
  \bot&\leftarrow\tat(\tram,\toffice)_k\wedge\tgo(\tram,\thome)_k\wedge\timet_{\kvar,\tmvar}\wedge\timet_{\kvar+1,\tmvar'}
        \quad\text{ for } d'-d \geq 16
\end{align}
and
$0\leq\kvar <\lambda-1$,
$0\leq\tmvar<\tmvar'\leq\tmflimit$.
For $\lambda=100$ and $\tmflimit=1000$,
this then amounts to roughly $10^8$ instances for each of the above constraints.

In what follows,
we characterize the effect of our formalization in terms of \HT-models,
and ultimately show the completeness and correctness of our translation.
\begin{definition}\label{def:ht:timed}
  An \HT\ interpretation \handt\
  over \alphabet\ with $\mathcal{T}\subseteq\alphabet$,
  is \emph{timed} wrt $\lambda\in\mathbb{N}$,

  if
  there is a timing function \tmf\ wrt $\lambda$
  such that
for all $0\leq\kvar<\lambda$,
  $\tmvar \in \mathbb{N}$,
  we have
  \begin{align*}
    \tuple{H,T}\models\timet_{\kvar,\tmvar}\text{ iff }\tmf(\kvar)=\tmvar
    \quad\text{ and }\quad
    \tuple{T,T}\models\timet_{\kvar,\tmvar}\text{ iff }\tmf(\kvar)=\tmvar
  \end{align*}
\end{definition}
We also call $\tmf$ the timing function induced by \handt.
\begin{proposition}\label{prop:ht:delta:timed}
  Let
  $\lambda,\tmflimit\in\mathbb{N}$.

  If $\tandt$ is an equilibrium model of $\Delta_{\lambda,\tmflimit}$
  then $\tandt$ is timed wrt $\lambda$.
\end{proposition}
\begin{proposition}\label{prop:ht:timed:delta}
  Let $\handt$ be an \HT\ interpretation
  and
  $\lambda\in\mathbb{N}$.

  If $\handt$ is timed wrt $\lambda$ and induces $\tmf$
  then $\handt\models\Delta_{\lambda,\tmflimit}$
  for $\tmflimit=\tmf(\lambda-1)$.
\end{proposition}
Clearly, the last proposition extends to equilibrium models.

Given a timed \HT-trace $\M = (\tuple{H_\kvar,T_\kvar}_{\rangeco{\kvar}{0}{\lambda}},\tmf)$ of length $\lambda$ over \alphabet,
we define $\mhtToHt{\M}$ as \HT\ interpretation
\(
\tuple{H\cup X ,T\cup X }
\)
where
\begin{align*}
  H &=\{a_\kvar\in\alphabetk \mid 0 \leq \kvar < \lambda, a\in H_\kvar \} &
  T &=\{a_\kvar\in\alphabetk \mid 0 \leq \kvar < \lambda, a\in T_\kvar \} \\
  X &=\{\timet_{\kvar, \tmvar} \mid \tmf(\kvar)=\tmvar, 0\leq\kvar<\lambda, \tmvar\in\mathbb{N}\}
\end{align*}
Note that $\mhtToHt{\M}$ is an \HT\ interpretation timed wrt $\lambda$.

Conversely,
given an \HT\ interpretation $\tuple{H,T}$ timed wrt $\lambda$ over $\alphabets\cup\alphabetT$
and its induced timing function $\tmf$,
we define $\htToMht{\tuple{H,T}}$ as
the timed \HT-trace
\begin{align*}
  (\tuple{ \{a\in\alphabet\mid a_\kvar\in H\}, \{a\in\alphabet\mid a_\kvar\in T\}}_{\rangeco{\kvar}{0}{\lambda}}, \tmf)
\end{align*}
In fact, both functions $\htToMhtf$ and $\mhtToHtf$ are invertibles,
and we get a one-to-one correspondence between \HT\ interpretations timed wrt $\lambda$ and timed \HT-traces of length $\lambda$.

Finally, we have the following completeness and correctness result.
\begin{theorem}[Completeness]\label{thm:mlp:ht:completeness}
  Let \program\ be a metric logic program
  and
  $\M=(\tuple{\Ttrace,\Ttrace}, \tmf)$ a total timed \HT-trace of length $\lambda$.

  If
  $\M$ is an metric equilibrium model of $\program$,
  then
  $\mhtToHt{\M}$ is an equilibrium model of $\Pi_\lambda(\program)\cup\Delta_{\lambda,\tmflimit}\cup\Psi_{\lambda,\tmflimit}(\program)$
  with $\tmflimit=\tmf(\lambda-1)$.
\end{theorem}
\begin{theorem}[Correctness]\label{thm:mlp:ht:correctness}
  Let
  \program\ be a metric logic program,
  and
  $\lambda,\tmflimit \in \mathbb{N}$.

  If
  $\tandt$ is an equilibrium model of $\Pi_\lambda(\program)\cup\Delta_{\lambda,\tmflimit}\cup\Psi_{\lambda,\tmflimit}(\program)$,
  then
  $\htToMht{\tandt}$ is a metric equilibrium model of $\program$.
\end{theorem}
 \section{Translation of Metric Logic Programs to \HTC}
\label{sec:mlp:htc}

We now present an alternative, refined formalization of the second and third parts,
utilizing integer variables and difference constraints to capture the timing function more effectively.
To this end,
we use the logic of \HTC\ to combine the Boolean nature of ASP with constraints on integer variables.

Given base alphabet $\alphabet$ and $\lambda\in\mathbb{N}$,
we consider the \HTC\ signature $\tuple{\mathcal{X},\mathcal{D},\mathcal{C}}$
where\footnote{In \HTC~\cite{cakaossc16a},
  Boolean variables are already captured by truth values \true\ and \undefined\ (rather than \false[alse]).}
\begin{align*}
  \mathcal{X} &= \alphabets \cup \{ \timet_k \mid 0\leq k<\lambda\} \\
  \mathcal{D} &= \{\true\}\cup\mathbb{N}\\
  \mathcal{C} &= \{a=\true\mid a\in\alphabets\}
                 \cup \{\x=\tmvar \mid \x    \in \mathcal{X}\setminus\alphabets,\tmvar\in \mathbb{N} \}\;\cup\\
              &\quad\;\{\x-\y\leq \tmvar \mid \x,\y \in \mathcal{X}\setminus\alphabets,\tmvar\in\mathbb{N} \}
\end{align*}
Rather than using Boolean variables,
this signature represents timing functions $\tau$ directly by integer variables $t_k$,
capturing that $\tau(k)=t_k$ for $0\leq k<\lambda$.
This is enforced by the integer constraints in $\mathcal{C}$,
whose meaning is defined by the following denotations:
\begin{align*}
  \den{a=\true}          &= \{v\in\mathbb{V}\mid v(a)=\true\} \qquad\qquad\qquad\text{ for all }a\in \alphabets\\
  \den{\x=\tmvar}        &= \{v\in\mathbb{V}\mid v(\x),       \tmvar\in\mathbb{N},\ v(\x)= \tmvar\}\\
  \den{\x-\y\leq \tmvar} &= \{v\in\mathbb{V}\mid v(\x),v(\y), \tmvar\in\mathbb{N},\ v(\x)-v(\y)\leq \tmvar\}
\end{align*}
This leads us to the the following counterpart of $\Delta_{\lambda,\tmflimit}$ in~\eqref{def:ht:delta}.
Given $\lambda \in \mathbb{N}$, we define
\begin{align}\label{def:htc:delta}
  \Delta^{c}_{\lambda} &= \{ \timet_{0} = 0 \} \cup \{ \timet_{\kvar}-\timet_{\kvar{+}1} \leq -1 \mid 0 \leq \kvar < \lambda-1 \}
\end{align}
Starting from $\timet_0=0$,
the difference constraints in~\eqref{def:htc:delta} enforce that $\timet_{\kvar}<\timet_{\kvar{+}1}$
reflecting that $\tau(0)=0$ and $\tmf(\kvar)<\tmf(\kvar+1)$ for $0\leq\kvar<\lambda-1$.
Moreover, $\Delta^{c}_{\lambda}$ is unbound and thus imposes no restriction on timing functions.
And no variable $\timet_{\kvar}$ is ever undefined:
\begin{proposition}\label{pro:htc:defined}
Let $\htcinterp$ be an \HTC\ interpretation and $\lambda \in \mathbb{N}$

If $\htcinterp\models\Delta^{c}_{\lambda}$, then $\Vh(\timet_{\kvar})\in\mathbb{N}$ for all $0\leq\kvar<\lambda$.
\end{proposition}
We also have $\Vt(\timet_{\kvar})\in\mathbb{N}$ by definition of \HTC\ interpretations, that is, since $\Vh\subseteq\Vt$.

Our variant of the third part of our translation re-expresses the ones in~(\ref{def:ht:psi:one}/\ref{def:ht:psi:two})
in terms of integer variables and difference constraints.
Given $\lambda \in \mathbb{N}$ and a metric logic program $\program$,
we define
\begin{align}
  \Psi^{c}_{\lambda}(\program) \eqdef
  &\ \{ \bot \leftarrow \tk{\body} \wedge \neg (\timet_{\kvar}-\timet_{\kvar+1} \leq -m) \mid
    0 \leq \kvar < \lambda-1,
    \label{def:htc:psi:one}\\&\qquad\qquad\nonumber
  \phantom{n\in\mathbb{N},} \melrule{\metric{\next}{m}{n} a \leftarrow \body} \in \program\} \; \cup \\
  &\ \{ \bot \leftarrow \tk{\body} \wedge \neg (\timet_{\kvar+1} - \timet_{\kvar} \leq n - 1) \mid
    0 \leq \kvar < \lambda-1,
    \label{def:htc:psi:two}\\&\qquad\qquad\nonumber
           n\in\mathbb{N},  \melrule{\metric{\next}{m}{n} a \leftarrow \body} \in \program\}
\end{align}

In fact, both $\Delta^{c}_{\lambda}$ and $\Psi^{c}_{\lambda}(\program)$ drop the upper bound on the range of a
timing function, as required in their Boolean counterparts.
Hence, their size is only proportional to $\mathcal{O}(\lambda)$,
and thus considerably smaller than their purely Boolean counterparts.

For the rule in~\eqref{ex:dentist:eit:translated},
we get
\begin{align}
  \bot&\leftarrow\tat(\tram,\toffice)_k\wedge\tgo(\tram,\thome)_k\wedge\neg (\timet_{\kvar}-\timet_{\kvar+1} \leq -15)
  \\
  \bot&\leftarrow\tat(\tram,\toffice)_k\wedge\tgo(\tram,\thome)_k\wedge\neg (\timet_{\kvar+1} - \timet_{\kvar} \leq 15)
\end{align}
for
$0 \leq \kvar < \lambda-1$.
Given $\lambda=100$, this only amounts to $10^2$ instances.

Mirroring our approach in Section~\ref{sec:mlp:ht},
we capture the meaning of $\Delta^{c}_{\lambda}$ using specialized \HTC-models.
This leads to the completeness and correctness of our translation.
\begin{definition}
  An \HTC\ interpretation \htcinterp\
  over $\tuple{\mathcal{X},\mathcal{D},\mathcal{C}}$
  is \emph{timed} wrt $\lambda$,
if
  there is a timing function \tmf\ wrt $\lambda$
  such that
  \(
  \Vh(\timet_{\kvar})=\tmf(\kvar)
  \)
  and
  \(
  \Vt(\timet_{\kvar})=\tmf(\kvar)
  \)
  for all $0\leq\kvar<\lambda$.
\end{definition}
As above, we call $\tmf$ the timing function induced by \htcinterp.

\begin{proposition}\label{prop:htc:delta:timed}
  Let $\htcinterp$ be an \HTC\ interpretation
  and
  $\lambda\in\mathbb{N}$.

  If $\htcinterp\models\Delta^{c}_{\lambda}$
  then $\htcinterp$ is timed wrt $\lambda$.
\end{proposition}
\begin{proposition}\label{prop:htc:timed:delta}
  Let $\htcinterp$ be an \HTC\ interpretation
  and
  $\lambda\in\mathbb{N}$.

  If $\htcinterp$ is timed wrt $\lambda$
  then $\htcinterp\models\Delta^{c}_{\lambda}$.
\end{proposition}
Unlike Proposition~\ref{prop:ht:delta:timed}, the latter refrain from requiring \HTC-interpretations in equilibrium.

Given an \HT-trace $\M = (\tuple{H_\kvar,T_\kvar}_{\rangeco{\kvar}{0}{\lambda}},\tmf)$ of length $\lambda$,
we define $\mhtToHtc{\M}$ as the \HTC\ interpretation
\(
\tuple{h\cup x,t\cup x}
\)
where $h,t,x$ are valuations such that
\begin{align*}
  h &=\{(a_\kvar,\true) \mid 0 \leq \kvar < \lambda, a\in H_\kvar \} &
  t &=\{(a_\kvar,\true) \mid 0 \leq \kvar < \lambda, a\in T_\kvar \} \\
  x &= \{(\timet_{\kvar},\tmvar) \mid \tmf(\kvar)=\tmvar, 0\leq\kvar<\lambda, \tmvar\in\mathbb{N} \}
\end{align*}
Similar to above, $\mhtToHtc{\M}$ is an \HTC\ interpretation timed wrt $\lambda$.

Conversely,
given an \HTC\ interpretation $\htcinterp$ timed wrt $\lambda$
and its induced timing function $\tmf$,
we define $\htcToMht{\htcinterp}$ as
the timed \HT-trace
\begin{align*}
  (\tuple{ \{a\mid\Vh(a_\kvar)=\true\}, \{a\mid\Vt(a_\kvar)=\true\}}_{\rangeco{\kvar}{0}{\lambda}}, \tmf)
\end{align*}
As above, functions $\htToMhtf^c$ and $\mhtToHtf^c$ are invertibles.
Hence,
we get a one-to-one correspondence between \HTC\ interpretations timed wrt $\lambda$ and timed \HT-traces of length $\lambda$.

Finally, we have the following completeness and correctness result.
\begin{theorem}[Completeness]\label{thm:mlp:htc:completeness}
  Let \program\ be a metric logic program
  and
  $\M=(\tuple{\Ttrace,\Ttrace}, \tmf)$ a total timed \HT-trace of length $\lambda$.

  If
  $\M$ is an metric equilibrium model of $\program$,
  then
  $\mhtToHtc{\M}$ is a constraint equilibrium model of $\Pi_\lambda(\program)\cup\Delta^c_{\lambda}\cup\Psi^c_{\lambda}(\program)$.
\end{theorem}
\begin{theorem}[Correctness]\label{thm:mlp:htc:correctness}
  Let
  \program\ be a metric logic program,
  and
  $\lambda \in \mathbb{N}$.

  If
  $\htcttinterp$ is a constraint equilibrium model of $\Pi_\lambda(\program)\cup\Delta^c_{\lambda}\cup\Psi^c_{\lambda}(\program)$,
  then
  $\htcToMht{\htcttinterp}$ is a metric equilibrium model of $\program$.
\end{theorem}

\section{Implementation}
\label{sec:implementation}

In what follows,
we rely on a basic acquaintance with the ASP system \clingo~\cite{PotasscoUserGuide}.
We outline specialized concepts as they are introduced throughout the text.
We show below how easily our approach is implemented via \clingo's meta encoding framework.
This serves us as a blueprint for a more sophisticated future implementation.
\Clingo\ allows for reifying a ground logic program in terms of facts, which can then be (re)interpreted by a meta encoding.
The result of another grounding is then channeled to the respective back-end, in our case a regular or hybrid ASP solver, respectively.
Though, for brevity, we must refer to~\cite{karoscwa21a} for details,
we mention that a reified ground program is represented by instances of predicates
\lstinline{atom_tuple},
\lstinline{literal_tuple},
\lstinline{rule},
\lstinline{output},
etc.
\footnote{Below we draw upon the symbol table, captured by \lstinline{output/2}, for extracting syntactic entities.}

\begin{lstlisting}[caption={Timed meta encoding (\texttt{meta.lp})},label={prg:meta},language=clingos]
time(0..lambda-1).%%#(\label{meta:time}#)

conjunction(B,T) :- literal_tuple(B), time(T),
       hold(L,T) : literal_tuple(B, L), L > 0;
   not hold(L,T) : literal_tuple(B,-L), L > 0.

body(normal(B),T) :- rule(_,normal(B)), conjunction(B,T).
body(sum(B,G),T)  :- rule(_,sum(B,G)), time(T),
  #sum{W,L :     hold(L,T), weighted_literal_tuple(B, L,W), L>0;
       W,L : not hold(L,T), weighted_literal_tuple(B,-L,W), L>0} >= G.

  hold(A,T) : atom_tuple(H,A)   :- rule(disjunction(H),B), body(B,T).
{ hold(A,T) : atom_tuple(H,A) } :- rule(     choice(H),B), body(B,T).
\end{lstlisting}
Listing~\ref{prg:meta} modifies the basic meta encoding in~\cite{karoscwa21a}
by adding a variable \lstinline{T} for time steps to all derived predicates.
Their range is fixed in Line~\ref{meta:time}.
In this way, an atom \lstinline{hold(a,k)} stands for $a_k$ in \alphabets,
where \lstinline{a} is the numeric identifier of $a$ in \alphabet.
While this encoding handles Boolean connectives,
the metric ones are treated in Listing~\ref{prg:metabound}.
The rules in Line~\ref{meta:bounded:true:one} and~\ref{meta:bounded:true:two} restore the symbolic representation
of the numerically identified atoms, which allows us to analyze the inner structure of modalized propositions.
Lines~(\ref{meta:bounded:initially:one}/\ref{meta:bounded:initially:two}) and
(\ref{meta:bounded:finally:one}/\ref{meta:bounded:finally:two})
deal with \initially\ and \finally, respectively.
Lines~\ref{meta:bounded:next:one} and~\ref{meta:bounded:next:two} realize the metric next operator,
$\metricI{\Next}a$, represented by term \lstinline{next(I,a)}.
Together Listing~\ref{prg:meta} and~\ref{prg:metabound} account for $\Pi_\lambda(\program)$.
\begin{lstlisting}[caption={Meta encoding for metric part of $\Pi_\lambda(\program)$ (\texttt{bounded.lp})},label={prg:metabound},language=clingos]
true(O,K) :- output(O,B), time(K), hold(L,K) : literal_tuple(B,L).%%#(\label{meta:bounded:true:one}#)
hold(L,K) :- output(O,B), time(K), true(O,K),  literal_tuple(B,L).%%#(\label{meta:bounded:true:two}#)

:- true(initially,K), time(K), K!=0.%%#(\label{meta:bounded:initially:one}#)
true(initially,0).%%#(\label{meta:bounded:initially:two}#)

:- true(finally,K), time(K), K!=lambda-1.%%#(\label{meta:bounded:finally:one}#)
true(finally,lambda-1).%%#(\label{meta:bounded:finally:two}#)

:- true(next(_,_),K), time(K), K=lambda-1.%%#(\label{meta:bounded:next:one}#)
true(V,K+1) :- true(next(I,V),K).%%#(\label{meta:bounded:next:two}#)
\end{lstlisting}

When expressing time via Boolean variables,
the two previous listings are combined with Listing~\ref{prg:time:ht} and~\ref{prg:interval:ht} below,
which realize $\Delta_{\lambda,\tmflimit}$ and $\Psi_{\lambda,\tmflimit}(\program)$, respectively.
Atoms $\timet_{\kvar,\tmvar}$ in \alphabetT\ are represented by \lstinline{t(k,d)}.
The upper bound $\tmflimit$ on the timing function's range is given by \lstinline{v},
and $\omega$ is represented by \lstinline{w}.
The two encodings directly mirror the definitions of $\Delta_{\lambda,\tmflimit}$ and
$\Psi_{\lambda,\tmflimit}(\program)$,
with one key difference:
the rule bodies in~\eqref{def:ht:psi:one} and~\eqref{def:ht:psi:two} are replaced in
Lines~\ref{interval:ht:one} and~\ref{interval:ht:two} of Listing~\ref{prg:interval:ht} by auxiliary atoms of
predicate \lstinline{true/2}.
\begin{lstlisting}[caption={Meta encoding for $\Delta_{\lambda,\tmflimit}$ (\texttt{time.lp})},label={prg:time:ht},language=clingos]
timepoint(0..v).

t(0,0).
t(K+1,D') : D<D', timepoint(D') :- t(K,D), time(K+1).%%#(\label{time:ht:two}#)
\end{lstlisting}
\begin{lstlisting}[caption={Meta encoding for $\Psi_{\lambda,\tmflimit}(\program)$ (\texttt{interval.lp})},label={prg:interval:ht},language=clingos]
:- true(next((M,N),V),K), t(K,D), t(K+1,D'), D' - D < M.%%#(\label{interval:ht:one}#)
:- true(next((M,N),V),K), t(K,D), t(K+1,D'), D' - D >= N , N!=w.%%#(\label{interval:ht:two}#)
\end{lstlisting}

When expressing time in terms of integer variables,
we rely on difference constraints for modeling timing functions.
Such simplified linear constraints have the form `$x - y \leq d$' for $x,y\in\mathcal{X}$ and $d\in\mathbb{Z}$
and are supported by the \clingo\ extensions \clingcon~\cite{bakaossc16a} and \clingodl~\cite{jakaosscscwa17a}.
We use below \clingcon's syntax and represent them as `\lstinline[mathescape]|&sum{$x$ ; $y$} <= $d$|'.
A Boolean atom $a$ can be seen as representing `$a=\true$'.
In the case at hand,
Listing~\ref{prg:meta} and~\ref{prg:metabound} are now completed by Listing~\ref{prg:time:htc} and~\ref{prg:interval:htc} below.
As above, they faithfully replicate the definitions of $\Delta^{c}_{\lambda}$ and $\Psi^{c}_{\lambda}(\program)$.
Unlike above, however,
the timing function is now captured by integer variables of form \lstinline{t(k)} and
its range restriction is now obsolete.
The rules in Listing~\ref{prg:time:htc} mirror the two conditions on timing functions,
namely, that \lstinline{t(0)} equals zero and that the instances of \lstinline{t(K+1)} receive a strictly greater integer
than the ones of \lstinline{t(K)} for \lstinline{K} ranging from \lstinline{0} to \lstinline{lambda-1}.
Similarly, given that \lstinline{w} stands for $\omega$,
the two rules in Listing~\ref{prg:interval:htc} correspond to
the difference constraints in~\eqref{def:htc:psi:one} and~\eqref{def:htc:psi:two}.
\begin{lstlisting}[caption={Meta encoding for $\Delta^{c}_{\lambda}$ (\texttt{time-c.lp})},label={prg:time:htc},language=clingos]
&sum{t(0)} = 0.
&sum{t(K) ; -t(K+1)} <= -1 :- time(K), time(K+1).
\end{lstlisting}
\begin{lstlisting}[caption={Meta encoding for $\Psi^{c}_{\lambda}(\program)$ (\texttt{interval-c.lp})},label={prg:interval:htc},language=clingos]
&sum{t(K); -t(K+1)} <= -M  :- true(next((M,N),V),K), time(K), time(K+1).
&sum{t(K+1); -t(K)} <= N-1 :- true(next((M,N),V),K), time(K), time(K+1), N!=w.
\end{lstlisting}
As in Listing~\ref{prg:interval:ht}, we use auxiliary atoms of predicate \lstinline{true/2} rather than the
corresponding rule bodies.
Notably, we shifted in Listing~\ref{prg:interval:htc} the difference constraints in~\eqref{def:htc:psi:one}
and~\eqref{def:htc:psi:two} from the body to the head.
This preserves (strong) equivalence in \HTC\ whenever all variables comprised in a constraint atom are defined,
as guaranteed by Proposition~\ref{pro:htc:defined}.

\begin{lstlisting}[caption={Metric logic program for dentist example (\texttt{dentist.lp})},label={prg:dentist},language=clingos]
item(icard). item(cash).
loc(dentist). loc(office). loc(atm). loc(home).

at(ram,office) :- initially.
at(cash,atm)   :- initially.
at(icard,home) :- initially.

go(ram,L') : loc(L'), L'!=L :- at(ram,L), not finally.

has(ram,I) :- at(ram,L), at(I,L), item(I).
at(ram,L)  :- at(ram,L), has(ram,I).

next((D,D+1),at(ram,L')) :- at(ram,L), go(ram,L'), distance(L,L',D).%%#(\label{ex:dentist:eit:code}#)

next((0,w),has(ram,I)) :- has(ram,I), not finally.
next((0,w),at(I,L)) :- at(I,L), item(I), not has(_,I), not finally.
\end{lstlisting}
Listing~\ref{prg:dentist} illustrates the above concepts using the dentist example
(leaving out the representation of Table~\ref{table:dentist} in terms of \lstinline{distance/3}).
For simplicity, we assume that each rule is implicitly in the scope of an always operator \alwaysF.
Moreover, included temporal operators and their comprised atoms are exempt from simplifications during grounding.\footnote{In meta encodings, this is done by adding corresponding \lstinline{#external} directives.}
We use predicate \lstinline{next/2} for the metric next operator.
As an example,
consider the instance of Line~\ref{ex:dentist:eit:code} for moving from $\toffice$ to $\thome$,
viz.\ the counterpart of~\eqref{ex:dentist:eit:translated}.
\begin{lstlisting}[language=clingos]
next((15,16),at(ram,home)) :-
             at(ram,office), go(ram,home), distance(office,home,15).
\end{lstlisting}
Note that \lstinline[mathescape]{next((0,w),$\cdot$)} in the head of the last two rules
stands for $\next$ aka $\next_{\intervco{0}{\omega}}$.

Note that the above encoding does not enforce that Ram is at the dentist in one hour with all necessary items.
Ideally,
this is represented with a global operator like $\metricI{\eventuallyF}\phi$.
In our example, this amounts to adding the (informal) rules
\begin{lstlisting}[mathescape,language=clingos]
:- initially, not $\eventuallyF_{\intervco{0}{61}}$goal
goal :- at(ram,dentist), has(ram,icard), has(ram,cash).
\end{lstlisting}
In the two approaches at hand,
we may compensate the lack of global metric operators by
replacing Line~1 by `\lstinline{:- finally, not goal}' and
enforcing the time limit either
by setting $\tmflimit$ to 60 in our \HT-based approach or
by extending Listing~\ref{prop:htc:timed:delta} with `\lstinline|&sum{t(K)} <= 60 :- time(K), not time(K+1).|' in our \HTC-based approach.
However, this is only effective when the goal is achieved at the final step.

The example in Listing~\ref{prg:dentist}
is addressed with \clingo\ in the following way.
\begin{lstlisting}[basicstyle=\small\ttfamily,numbers=none]
clingo dentist.lp --output=reify |
clingo 0 - meta.lp bounded.lp time.lp interval.lp -c lambda=4 -c v=110
\end{lstlisting}
When using \clingcon\ instead,
it suffices to replace the second line with:
\begin{lstlisting}[basicstyle=\small\ttfamily,numbers=none]
clingcon 0 - meta.lp bounded.lp time-c.lp interval-c.lp -c lambda=4
\end{lstlisting}
Our choices of \lstinline{lambda} and \lstinline{v}
allow for all movement combinations within the 4 steps required to reach the goal;
we obtain in each case 27 solutions.
Once we include the query-oriented additions from above, we obtain a single model instead.

To gather some indicative results on the scalability of both approaches,
we multiplied the durations in Table~\ref{table:dentist} (and limit \lstinline{v} for \clingo) with 1, 5, and 10
and summarize the results in Table~\ref{tab:benchmarks}.
Despite the rather limited setting,
we observe that the usage of integer variables leads to a complete independence
of performance and time granularity.
\begin{table}[ht]
  \centering
\begin{tabular}{|r|rrr|rrr|rrr|}
      \hline
         &\multicolumn{3}{c|}{\clingo~\cite{karoscwa21a}}
         &\multicolumn{3}{c|}{\clingcon~\cite{bakaossc16a}}
         &\multicolumn{3}{c|}{\clingodl~\cite{jakaosscscwa17a}}\\
         & solve& ground&\#rules &solve&ground&\#rules&solve&ground&\#rules\\
      \hline
       1 &    0.01&  0.281&  297211& 0.00& 0.020&  2165 & 0.00& 0.018&  2165 \\
       5 &    7.09& 19.358& 7527451& 0.00& 0.020&  2165 & 0.00& 0.018&  2165 \\
      10 & 609.46s&143.083&30177751& 0.00& 0.020&  2165 & 0.00& 0.018&  2165 \\
    \hline
    \end{tabular}
    \medskip
    \caption{Indicative results for \clingo, \clingcon, \clingodl\ (times in seconds; no resource limits).}
    \label{tab:benchmarks}
\end{table}  \section{Conclusion}\label{sec:discussion}

We presented a computational approach to metric ASP that allows for fine-grained timing constraints.
We developed two alternative translations from firm semantic foundations, and proved their completeness and correctness.
Our second translation has a clear edge over the first one, when it comes to a fine-grained resolution of time.
This is achieved by outsourcing the treatment of time.
Clearly, this is not for free either.
\clingodl\ maps difference constraints into graphs, whose nodes are time variables and weighted edges reflect the actual
constraints.
This results in a quadratic space complexity.
\clingcon\ pursues a lazy approach to constraint solving that gradually unfolds an ASP encoding of linear constraints.
In the worst case, this amounts to the space requirements of our first translation.
As well, such constraints are hidden from the ASP solver and cannot be used for directing the search.
Hence,
despite our indicative observations,
a detailed empirical analysis is needed to account for the subtleties of our translation and its target systems.

However, a prominent use case involves employing the identity (timing) function,
where intervals reference only state indices within traces.
This coarser notion of time reduces the discrepancy between our two translations.
Further improvement is possible through more sophisticated Boolean encodings, such as an order
encoding~\cite{crabak94a,bageinospescsotawe15a}.

 \bibliographystyle{./include/latex-class-llncs/splncs04}

 \end{document}